\documentclass[10pt, a4paper, conference, compsocconf]{IEEEtran}

\usepackage{cite}
\usepackage{graphicx}
\usepackage{amsmath}
\usepackage{amssymb}

\newcommand{\citet}[1]{\cite{#1}}

\DeclareGraphicsExtensions{.pdf,.jpeg,.jpg,.png}

\DeclareMathOperator*{\argmax}{arg\,max}

\usepackage{hyperref}
\hypersetup{
	bookmarks=true,
	pdftoolbar=true,
	pdfmenubar=true,
	pdffitwindow=false,
    pdftitle={A Person Re-Identification System For Mobile Devices},
    pdfauthor={George Cushen},
	pdfcreator={},
	pdfproducer={},
    pdfnewwindow=true,
    colorlinks=false,
    linkcolor=red,
    citecolor=green,
    filecolor=magenta,
    urlcolor=cyan
}

\usepackage{tikz}
\newcommand\copyrighttext{
	\footnotesize \textcopyright 2015 IEEE. Personal use of this material is permitted. Permission from IEEE must be obtained for all other users, including reprinting/republishing this material for advertising or promotional purposes, creating new collective works for resale or redistribution to servers or lists, or reuse of any copyrighted components of this work in other works.}
\newcommand\copyrightnotice{
	\begin{tikzpicture}[remember picture,overlay]
	\node[anchor=south,yshift=10pt] at (current page.south) {\fbox{\parbox{\dimexpr\textwidth-\fboxsep-\fboxrule\relax}{\copyrighttext}}};
	\end{tikzpicture}
}

\usepackage{url}

\begin{document}

\title{A Person Re-Identification System For Mobile Devices}

\author{\IEEEauthorblockN{George A.\ Cushen}
\IEEEauthorblockA{University of Southampton, UK\\
gc505@ecs.soton.ac.uk
}
}

\maketitle
\copyrightnotice
\begin{abstract}
Person re-identification is a critical security task for recognizing a person across spatially disjoint sensors. Previous work can be computationally intensive and is mainly based on low-level cues extracted from RGB data and implemented on a PC for a fixed sensor network (such as traditional CCTV). We present a practical and efficient framework for mobile devices (such as smart phones and robots) where high-level semantic soft biometrics are extracted from RGB and depth data. By combining these cues, our approach attempts to provide robustness to noise, illumination, and minor variations in clothing. This mobile approach may be particularly useful for the identification of persons in areas ill-served by fixed sensors or for tasks where the sensor position and direction need to dynamically adapt to a target. Results on the BIWI dataset are preliminary but encouraging. Further evaluation and demonstration of the system will be available on our website.
\end{abstract}

\begin{IEEEkeywords}
person re-identification; mobile; soft biometrics; clothing recognition;
\end{IEEEkeywords}

\section{Introduction}

Person re-identification is a critical security task for recognising a person across spatially disjoint sensors. Besides the rapidly growing number of published papers on person re-identification, the importance of this field is recognised in the recent survey by \citet{Vezzani2014People}, published in a book on `Person Re-Identification'~\citet{Gong2014Person}. The identification problem may be classed as either \textit{single shot} or \textit{multi-shot}. Single shot  \cite{Koestinger2012Large, Avraham2012Learning, Zhao2013Unsupervised} approaches are only able to utilize one image of each person whereas multi-shot \cite{Farenzena2010Persona, Salvagnini2013Person} approaches can exploit multiple images of each person. The related work generally establishes either new feature representations \cite{Farenzena2010Persona, Zhao2013Unsupervised} or discriminative matching models \cite{Koestinger2012Large, Avraham2012Learning}. The closest work to ours is perhaps that of \cite{Yang2011Real-time} and \cite{Layne2014Investigating}. The approach of \cite{Yang2011Real-time} is limited to tagging 8 attributes on a simple dataset in a standard lab environment. Whereas Layne et al.\ \cite{Layne2014Investigating} discuss the new challenges associated with mobile re-identification. Their work is the state of the art and has only recently been published at the time of writing.

\begin{figure*}
	\centering
	\includegraphics[width=1\textwidth]{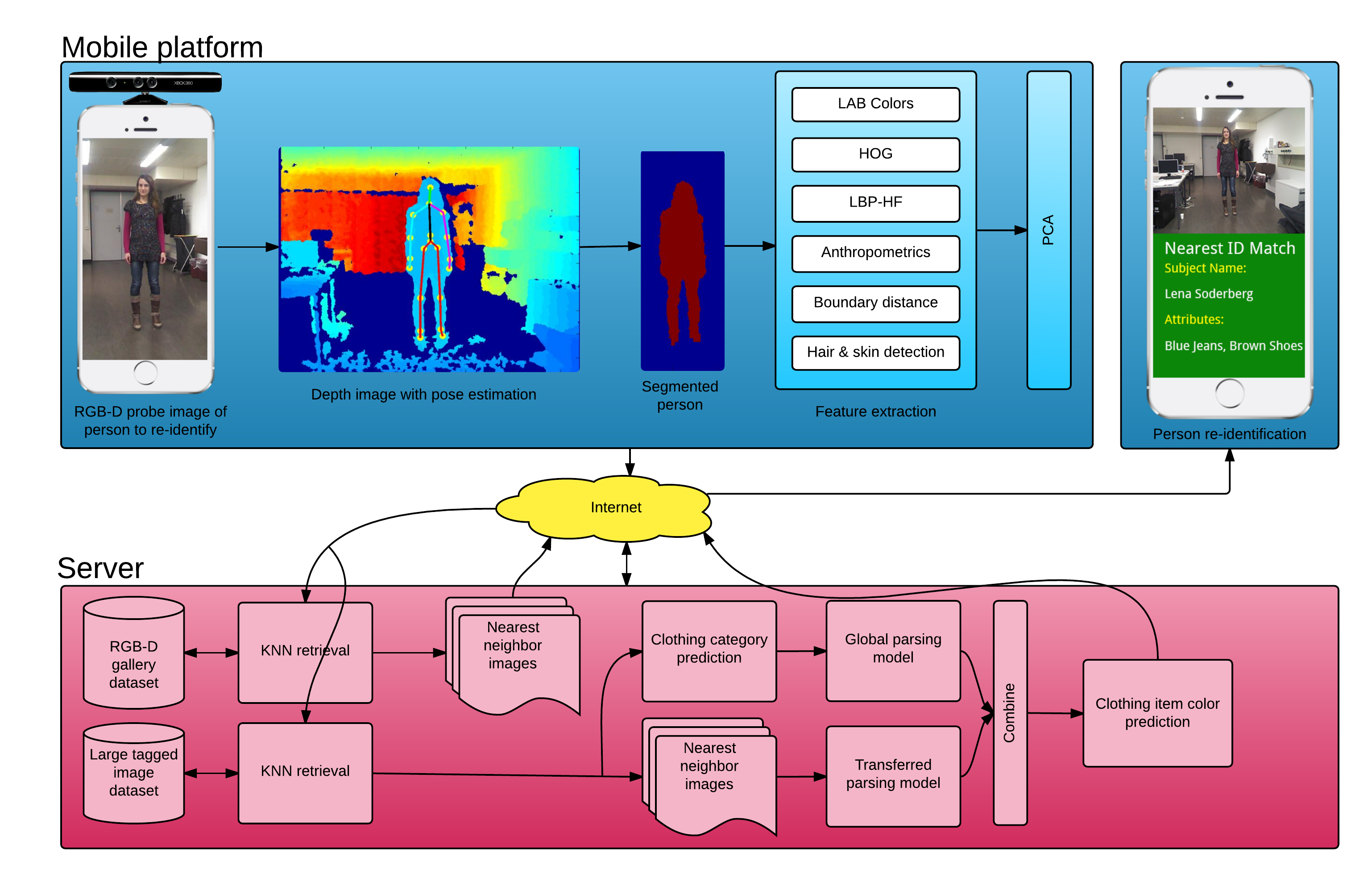}	
	\caption{Overview of our mobile re-identification pipeline.}\label{fig:intro:sys:bio}	
\end{figure*}

Person identification has been an active area of research for the past decade, but has almost exclusively focused on popular 2D RGB image data captured from fixed positions. This is a logical place to start since many venues already have a large network of traditional surveillance cameras which produce RGB data. Recently, more advanced sensor types such as the Kinect have become very popular and available at a low cost. Smart mobile devices have also become very popular and have been considered in related work for clothing retrieval~\cite{Cushen2013Mobile}. Gartner is predicting $1.37$ billion smart phone sales and $320$ million tablet sales for 2015. In terms of mobile OS, Android is expected to lead with $53\%$ of the market~\cite{GartnerInc2015Worldwide}. The diversity of mobile platforms with integrated sensors is growing with new technologies such as wearable devices (Google Glass), intelligent robots, and remotely operated vehicles. Given these observations, we believe it is time to extend the literature to new up-and-coming scenarios. This is the focus of this work, where we present a novel semantic approach that integrates RGB and depth data to extract clothing and skeletal soft biometrics in a re-identification system for mobile devices. An overview of the system is depicted in Figure \ref{fig:intro:sys:bio}. Since mobile devices have very limited computing resources available, much attention is given to efficiency unlike most related work which is computationally intensive and runs on powerful workstations. This mobile approach may be particularly useful for the identification of persons in areas ill-served by fixed sensors or for tasks where the sensor position and direction need to dynamically adapt to a target. Furthermore, we contribute semantic ground truth clothing labels for the \textit{BIWI} dataset to enable evaluation of predicted clothing items.

This paper is structured as follows: section 2 discusses the datasets used. Section 3 describes the feature descriptors and retrieval process. In section 4 the semantic soft biometric clothing recognition method is described. Section 5 covers the preliminary results. Finally, section 6 concludes the paper.

\section{Datasets}
\label{sec:datasets}

Most RGB-D datasets of people are targeted towards activity and gesture recognition. Very few RGB-D datasets exist for the purpose of person re-identification. In \cite{Barbosa2012Re-identification}, the first dataset explicitly for the purpose of RGB-D re-identification is created but there are few frames available per subject and the faces are blurred for privacy. We consider the state of the art \textit{BIWI} dataset~\cite{Munaro2014One-shot} which overcomes these limitations. 

The BIWI dataset is targeted to long-term people re-identification from RGB-D cameras. It contains 50 training and 56 testing sequences of 50 different people captured with a Microsoft Kinect for Windows at approximately 10fps. 28 of the people present in the training set have also been recorded in two testing videos each: \textit{still} and \textit{walking}. These were collected on a different day and in a different location with respect to the training dataset, so most people are dressed differently. Thus, this dataset can provide a challenge for re-identification.

For training the semantic clothing prediction, the annotated subset of the \textit{Fashionista} dataset~\cite{Yamaguchi2012Parsing} and the \textit{Paper Doll} dataset~\cite{Yamaguchi2013Paper} are utilized since they contain a significantly more diverse range of clothing than that present in the BIWI training dataset so can be applicable to many scenarios. The Fashionista dataset consists of 685 real-world photos from a popular online fashion social network, \href{http://chictopia.com}{chictopia.com}. For each of these photos, there are ground truth annotations consisting of 53 different clothing labels, plus hair, skin, and null labels. Whereas the Paper Doll dataset is a large collection of over $300,000$ fashion photos that are weakly annotated with clothing items, also collected from \href{http://chictopia.com}{chictopia.com}.

In order to evaluate predicted clothing items worn by a subject in a given test sequence, ground truth clothing labels are required. As these are not available in the BIWI dataset, we contribute semantic ground truth clothing labels which will be published online (see URL in section \ref{sec:conc}). First, five crowd-sourcing users are chosen to manually identify the clothing present. They are shown each image sequence in the BIWI dataset and given a choice of clothing attributes from the Fashionista dataset to tag. Three or more votes are required for a tag to become associated with the sequence. To ensure high quality annotations, the received annotations are verified.

\section{Mobile Re-Identification}

Our client-server framework for mobile devices identifies a subject facing a depth camera given a single frame as input. In order to achieve this goal, we consider two different approaches. First, a clothing descriptor is computed and secondly, a skeletal descriptor is computed from the pose estimation provided by the Microsoft Kinect SDK. These two kinds of soft biometrics have been chosen since they are relatively efficient to compute, which is important in a mobile framework, whilst also showing reasonable performance in previous work.

Person re-identification on mobile devices can differ significantly from the traditional re-identification environment of two or more fixed sensors. Consider the case of a security officer with wearable technology (similar to Google Glass). The mobile infrastructure should consistently detect and identify persons in the officer's field of view using only one sensor, but the target may enter and exit the view multiple times depending on the motion of each party. In this case, there is no longer the concept of separate probe and gallery datasets. An alternative mobile re-identification scenario is where a suspicious person was previously identified and recorded in a probe dataset. Mobile re-identification could help a robot or remotely operated vehicle to locate the subject from the observed sensor data. We focus more on the latter case in terms of probe and gallery datasets.

In this section, the pre-processing and features for person re-identification are described. First, a mobile device captures an image of the subject using a depth sensor. The Microsoft Kinect SDK is used to provide pose estimation and segmentation of the person from the RGB-D data, since the SDK is available and optimized for this purpose. Microsoft's tracking algorithm can only accurately estimate frontal poses because it is based on a classifier which has only been trained with frontal poses of people. Hence, we discard frames where any joint is reported by the SDK as untracked or where a face cannot be detected by Viola-Jones. 

A local clothing feature vector is calculated for each body part in the pose estimation based on Lab colour, LBP-HF, HOG, skin/hair detection and boundary distance (refer to subsection \ref{chap:bio:features}). These features are normalized and mean-std pooling of the features is calculated on a $4 \times 4$ grid. Then pooled feature vectors are concatenated into one representative clothing descriptor and PCA is performed to reduce dimensionality for efficient retrieval. Additionally, the anthropometric descriptor is computed as described below and concatenated for the purpose of re-identification.

\subsection{Features}
\label{chap:bio:features}

In this section, a discussion of the features employed in this approach is presented. 

The texture and shape are described by rotation-invariant local binary pattern histogram Fourier (LBP-HF) features since they have been shown to achieve very good general performance in an efficient manor\cite{Zhao2012Rotation-invariant}. The HOG descriptor provides further information about the shape\cite{Dalal2005Histograms}. The boundary distance is given by the negative log distance from the image boundary. The pose distance is given by the negative log distance from the pose estimation joints. A Lab colour descriptor is employed rather than one based on a different colour space such as RGB since it models the human vision system and is more perceptually uniform. Perceptually uniform means that a change in colour value should produce a change of about the same magnitude of visual importance.

Skin/hair detection gives the likelihood of skin or hair at the pixel. Generalized logistic regression is used to compute the likelihood based on Lab, LBP-HF, HOG, boundary distance, and pose distance for input. It is learned from the Fashionista dataset using a one-vs-all strategy.

Skeleton based descriptors yield a signature vector for a given subject based on their anthropometrics and body pose. Generally, skeletal trackers with the highest performance are those that work on 3D data, as opposed to 2D. However, feature-wise, 2D descriptors can perform better than their 3D counterparts~\cite{Nanni2015Ensemble}. Therefore, we let the Kinect tracker locate skeleton joints in the 3D domain and then re-project them onto the 2D image plane by taking into account the sensor calibration. Once the joints are available in the 2D image domain, the skeleton features can be calculated.

\begin{figure}[t!]
	\centering
	\includegraphics[width=0.45\textwidth]{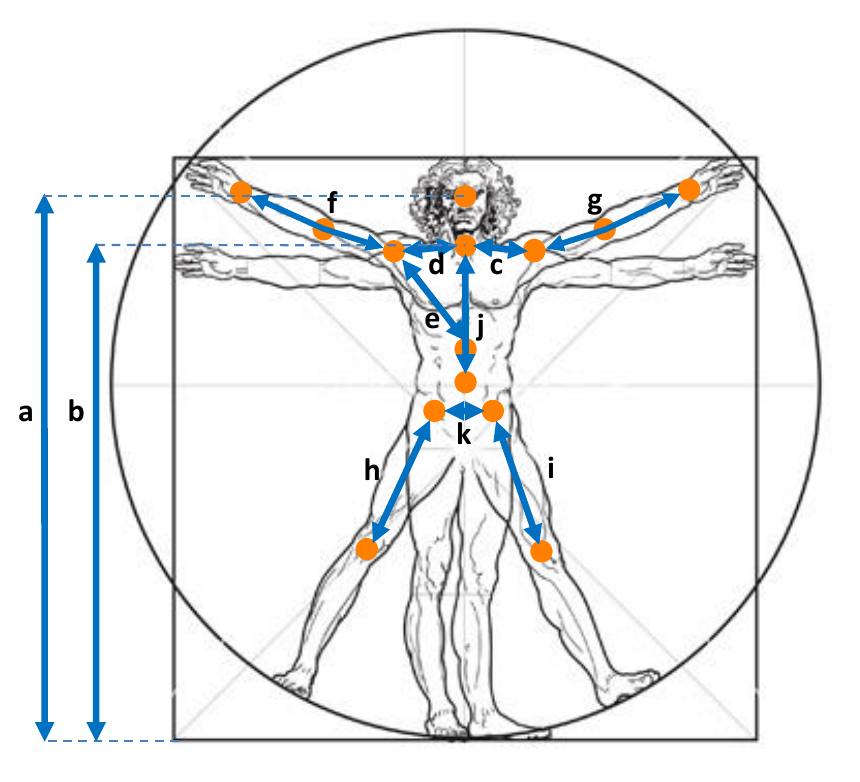}
	\caption{Distances and ratios utilized for the soft biometric anthropometric descriptor. Image courtesy of \cite{Munaro2014One-shot}.}
	\label{fig:bio:distances}
\end{figure}

We extract the following 13 skeleton features based on the work of \cite{Munaro2014One-shot}: a) head height, b) neck height, c) neck to left shoulder distance, d) neck to right shoulder distance, e) torso to right shoulder distance, f) right arm length, g) left arm length, h) right upper leg length, i) left upper leg length, j) torso length, k) right hip to left hip distance, l) ratio between torso length and right upper leg length (j/h), m) ratio between torso length and left upper leg length (j/i). These labels (a)-(m) correspond to those depicted in Figure \ref{fig:bio:distances}. The 13 features are then normalized and concatenated to form the skeleton descriptor.

It is assumed that the re-identification system is applied in an indoor scenario or outdoors during summertime when people often wear just one layer of clothing and not large heavy clothes that may occlude key feature points and distort the estimated anthropometrics.

\subsection{Retrieval}

In our approach, there are two retrieval algorithms. The former is used for retrieving identification results and the latter is used for predicting semantic attributes.

In a mobile infrastructure, there is a need for efficient identification and retrieval of subjects based on matching the feature vectors between the probe and each subject that is enrolled in the gallery database. For this purpose, the $L_2$ distance is minimized over the descriptors to obtain the K nearest neighbours (KNN) in the BIWI dataset. A KD-tree is constructed to efficiently index the samples.

The second retrieval algorithm is similar but operates on the Paper Doll dataset to retrieve similar clothing attributes to those present in the query image. It only considers the clothing descriptor as input and not the combined clothing and anthropometric descriptor like that used in the former retrieval algorithm.

Note that some retrieval precision could be sacrificed for increased speed by using a more approximate algorithm. This trade-off is explored in \cite{Muja2009Fast}. It is shown that with approximations, a speed increase of up to 3 orders of magnitude can be achieved over linear search (KD-trees) if we are willing to accept a lower precision and hence less neighbors returned are exact nearest neighbors. A significant decrease in precision would be unacceptable for re-identification purposes.

\section{Clothing Parsing}

In this section, an approach is described based on the work of \citet{Yamaguchi2013Paper} for detecting clothing attributes and localizing clothing items on the query image to enable semantic colour prediction of each clothing item. 

Let $y_i$ be the semantic label of the clothing item at pixel $i$ in the image. After clothing attributes have been predicted by the retrieval stage, the algorithm begins to parse the clothing in the query by computing the clothing likelihood $S$ of assigning clothing label $l$ to $y_i$ at pixel level by combining global $S_{global}$ and transfer $S_{transfer}$ models. This likelihood function $S$ is modeled as:
\begin{equation} \label{eq:bio:parsing}
\begin{split}
S(y_i \mid \mathbf{x}_i, D) \equiv & S_{global} \left(y_i \mid \mathbf{x}_i, D\right)^{\lambda_1} \cdot \\& S_{transfer} \left(y_i \mid \mathbf{x}_i, D\right)^{\lambda_2}
\end{split}
\end{equation}

where $\mathbf{x}_i$ denotes the features at pixel $i$, $\Lambda \equiv \left[ \lambda_1 , \lambda_2 \right]$ are weighting parameters. Since the gallery (retrieval) dataset that we utilize in this paper has a limited range of clothing items, we introduce a large dataset of tagged fashion images, Paper Doll, for predicting attributes present in the query image. Therefore, we let $D$ be the set of nearest-neighbours retrieved from the Paper Doll dataset.

\subsection{Global Parsing}

Global clothing likelihood is the first term in the clothing parsing model. It is modeled
as a logistic regression that computes a likelihood of a label assignment to each
pixel for a given set of possible clothing items:
\begin{equation}
S_{global} \left(y_i \mid \mathbf{x}_i, D\right) \equiv 
P \left(y_i = l \mid \mathbf{x}_i, \theta^g_l \right) \cdot \mathbf{1} [ l \in \tau(D) ]
\end{equation}

where $P$ is a logistic regression based on the feature vector $\mathbf{x}_i$ and model parameter $\theta^g_l$. Let $\tau(D)$ be a set of predicted clothing attributes given by the Paper Doll nearest-neighbour retrieval stage. Finally, let $\mathbf{1} [\cdots]$ be an indicator function defined as:
\begin{equation}
\mathbf{1}[l \in \tau(D)] = \mathbf{1}_{\tau(D)}(l) =
\begin{cases} 
1 &\text{if } l \in \tau(D), \\
0 &\text{if } l \notin \tau(D).
\end{cases}
\end{equation}

The following features are calculated for $\mathbf{x}_i$ in the logistic regression: Lab colour, pose distances, LBP-HF, and HOG. Note that unpredicted items are set a probability of $0$. The model parameter $\theta^g_l$ is trained on all the clothing items in the annotated training subset of the Fashionista dataset. 

\subsection{Transferred Parsing}

The transferred parse is the second stage of the parsing model. The mask likelihoods that were estimated by the global parse $S_{global}$ are transferred from the retrieved Paper Doll images to the query image.

First we compute an over-segmentation \cite{Felzenszwalb2004Efficient} of both the query and retrieved images. For each super-pixel in the query image, we find the nearest
super-pixels in each retrieved image using the aforementioned L2 pose distance and compute a concatenation of bag of words (BoW) from Lab, LBP-HF, and gradient features. The closest super-pixel from each retrieved image is chosen by minimizing the L2 distance on the BoW feature.

Let the transfer model be defined as:
\begin{equation}
S_{\mathit{transfer}} \left(y_i \mid \mathbf{x}_i, D\right) \equiv 
\frac{1}{Z} \sum_{r \in D} \frac{M(y_i,s_{i,r})}{1 + \| h(s_i) - h(s_{i,r}) \|}
\end{equation}

where $s_i$ is the super-pixel of pixel $i$, $s_{i,r}$ is the corresponding super-pixel from image $r$, $h(s)$ is the BoW features of super-pixel $s$, and $Z$ is a normalization constant. Additionally, let us denote $M$ as the mean of the global parse over super-pixel $s_{i,r}$:
\begin{equation}
\begin{split}
M&(y_i,s_i,r) \equiv \\
&\frac{1}{|s_{i,r}|} \sum_{j \in s_{i,r}} P \left(y_i = l \mid \mathbf{x}_i, \theta^g_l \right) \cdot \mathbf{1} [ l \in \tau(r) ]
\end{split}
\end{equation}
where $\tau(r)$is the set of clothing attributes for image $r$.

\subsection{Overall Likelihood}

Once the two likelihood terms have been computed, the final pixel likelihood $S$ is given by the previously defined equation \ref{eq:bio:parsing}.

However, there is the problem of choosing the weighting parameters $\Lambda$. The  weights are chosen such that an optimal foreground accuracy is achieved from the MAP assignment of pixel labels in the Fashionista training subset:

\begin{equation}
\max_\Lambda \sum_{i \in F}  \mathbf{1} \left[ \tilde{y}_i = \argmax_{y_i} S_\Lambda (y_i \mid \mathbf{x}_i) \right]
\end{equation}

where $F$ are the pixels in the foreground and $\tilde{y}_i$ is the ground-truth annotation of pixel $i$. The optimization is implemented with a simplex search algorithm.

\subsection{Semantic Clothing Color}

A final stage is presented to assign soft biometric clothing color attributes. These are important to enable natural language searching.

The English language contains eleven main color terms: `black', `white', `red', `green', `yellow', `blue', `brown', `orange', `pink', `purple', and `gray'. The estimated clothing color attribute will be chosen from this set. 

A color histogram is computed on each localized clothing item detected in the previous stage. In order to convert this numeric color space representation of clothing color to one that is more meaningful to humans, the X11 color names~\cite{Wikipedia2015X11} are considered. X11 is part of the X Window System which is commonly found on UNIX like systems such as the popular Ubuntu operating system. X11 colour names are represented in a file which maps certain color strings to color values. The mappings for the eleven main colors mentioned above are extracted from this file, allowing the dominant clothing color value to be matched to the closest value in the list, giving a semantic clothing color attribute for each item of clothing detected in the image.

\section{Experimental Results}

\begin{figure*}
	\centering	
	\includegraphics[width=0.9\textwidth]{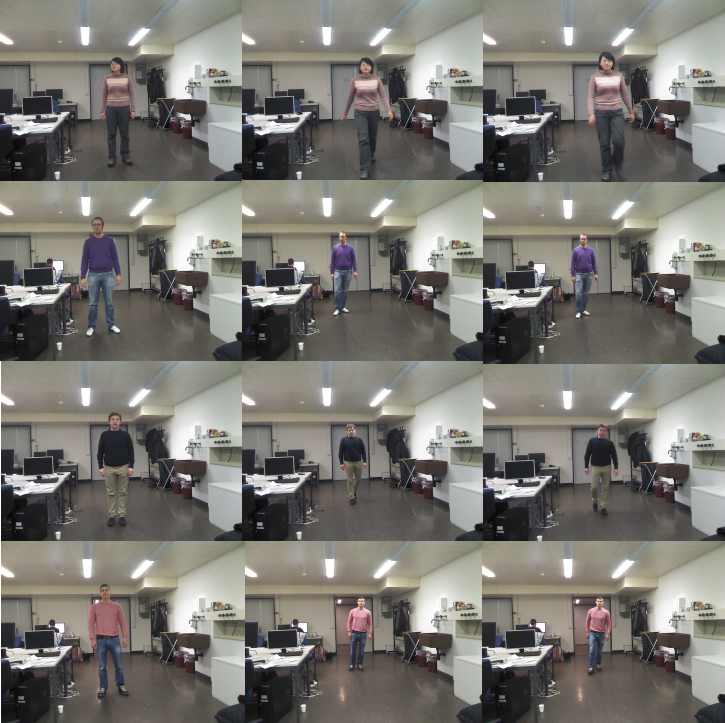}	
	\caption{Preliminary results on the BIWI dataset. First column shows RGB part of the RGB-D image for the subject (probe). Next two columns show retrieval results for subject re-identification.}\label{fig:bio:early-results}	
\end{figure*}

In the first experiment, the aim is to qualitatively test short-term person re-identification between the BIWI \textit{still} and \textit{walking} datasets. Although the results are preliminary, our system is capable of operating near real-time and we can see in figure \ref{fig:bio:early-results} that the results are encouraging as the persons are generally correctly re-identified between the \textit{still} and \textit{walking} datasets, implying a favorable rank-1. In some cases, clothing labels which are not observed in the probe image can rank highly - this may in part be due to the similarity of anthropometric features that are used in the retrieval alongside visual cues, and additional optimisation may be required. 

Considering related work, the approach in \cite{Yang2011Real-time} is faster than ours, but is less practical as it is limited to tagging 8 attributes on a simple dataset in a  lab environment without consideration of mobile scenarios or RGB-D. 

Further results along with a quantitative comparison with related work will be published on the author's website (see section \ref{sec:conc}).

The framework consists of the mobile client, which may be a smart phone or robot, and a server. The devices are wirelessly connected over a network such as the internet. In this work, we implement on a smart phone since these are currently the most popular kind of mobile device. Mobile consumer devices with depth sensors are yet to be made available, so we do not capture any input data on the mobile device for testing or demonstration and instead use the pre-recorded RGB-D data from the BIWI dataset as input. The client side is implemented with the Android SDK and NDK to achieve native processing speed and demonstrated on the Samsung Note 4 (1.3GHz Exynos 5433). The server side is implemented in Matlab, Python and C++, running on a quad-core 3.5GHz CPU. 

\section{Conclusions}
\label{sec:conc}
In this paper a person re-identification framework for mobile devices (such as smart phones and robots) is presented and the BIWI dataset is extended with ground truth clothing labels. For the case of smart phones and tablets, we assume that a depth sensor will be integrated into consumer mobile devices in the near future and we currently demonstrate with a pre-recorded input. The mobile device extracts clothing and skeletal features, reducing dimensions by PCA. The features are transmitted over the internet to a server which computes K-nearest neighbours to retrieve the closest matches from persons enrolled in the database and predicts semantic clothing attributes. Predicted clothing labels provide a meaningful soft biometric and can be useful to enable natural language based person searching or to yield a meaningful semantic description even when the subject has not been previously enrolled in the database. 

The results presented so far are preliminary but encouraging. Further implementation details, evaluation, and demonstration of the system will be available at \url{http://cushen.me}.

\bibliographystyle{IEEEtran}
\bibliography{cushen_sitis2015}

% Generated by IEEEtran.bst, version: 1.13 (2008/09/30)
\begin{thebibliography}{10}
\providecommand{\url}[1]{#1}
\csname url@samestyle\endcsname
\providecommand{\newblock}{\relax}
\providecommand{\bibinfo}[2]{#2}
\providecommand{\BIBentrySTDinterwordspacing}{\spaceskip=0pt\relax}
\providecommand{\BIBentryALTinterwordstretchfactor}{4}
\providecommand{\BIBentryALTinterwordspacing}{\spaceskip=\fontdimen2\font plus
\BIBentryALTinterwordstretchfactor\fontdimen3\font minus
  \fontdimen4\font\relax}
\providecommand{\BIBforeignlanguage}[2]{{%
\expandafter\ifx\csname l@#1\endcsname\relax
\typeout{** WARNING: IEEEtran.bst: No hyphenation pattern has been}%
\typeout{** loaded for the language `#1'. Using the pattern for}%
\typeout{** the default language instead.}%
\else
\language=\csname l@#1\endcsname
\fi
#2}}
\providecommand{\BIBdecl}{\relax}
\BIBdecl

\bibitem{Vezzani2014People}
R.~Vezzani, D.~{Baltieri}, and R.~{Cucchiara}, ``People reidentification in
  surveillance and forensics: {A} survey,'' \emph{{ACM} {Computing} {Surveys}
  ({CSUR})}, vol.~46, no.~2, p.~29, 2014.

\bibitem{Gong2014Person}
S.~Gong, M.~{Cristani}, and S.~{Yan}, \emph{Person {Re}-{Identification}
  ({Advances} in {Computer} {Vision} and {Pattern} {Recognition})}.\hskip 1em
  plus 0.5em minus 0.4em\relax Springer, Jan. 2014.

\bibitem{Koestinger2012Large}
M.~Koestinger, M.~{Hirzer}, P.~{Wohlhart}, P.~M. {Roth}, and H.~{Bischof},
  ``Large scale metric learning from equivalence constraints,'' in
  \emph{Computer {Vision} and {Pattern} {Recognition} ({CVPR}), 2012 {IEEE}
  {Conference} on}.\hskip 1em plus 0.5em minus 0.4em\relax {IEEE}, 2012, pp.
  2288--2295.

\bibitem{Avraham2012Learning}
T.~Avraham, I.~{Gurvich}, M.~{Lindenbaum}, and S.~{Markovitch}, ``Learning
  implicit transfer for person re-identification,'' in \emph{Computer
  {Vision}{\textendash}{ECCV} 2012. {Workshops} and {Demonstrations}}.\hskip
  1em plus 0.5em minus 0.4em\relax Springer, 2012, pp. 381--390.

\bibitem{Zhao2013Unsupervised}
R.~Zhao, W.~{Ouyang}, and X.~{Wang}, ``Unsupervised salience learning for
  person re-identification,'' in \emph{Computer {Vision} and {Pattern}
  {Recognition} ({CVPR}), 2013 {IEEE} {Conference} on}.\hskip 1em plus 0.5em
  minus 0.4em\relax {IEEE}, 2013, pp. 3586--3593.

\bibitem{Farenzena2010Persona}
M.~Farenzena, L.~{Bazzani}, A.~{Perina}, V.~{Murino}, and M.~{Cristani},
  ``Person re-identification by symmetry-driven accumulation of local
  features,'' in \emph{Computer {Vision} and {Pattern} {Recognition} ({CVPR}),
  2010 {IEEE} {Conference} on}.\hskip 1em plus 0.5em minus 0.4em\relax {IEEE},
  2010, pp. 2360--2367.

\bibitem{Salvagnini2013Person}
P.~Salvagnini, L.~{Bazzani}, M.~{Cristani}, and V.~{Murino}, ``Person
  re-identification with a ptz camera: an introductory study,'' in \emph{Image
  {Processing} ({ICIP}), 2013 20th {IEEE} {International} {Conference}
  on}.\hskip 1em plus 0.5em minus 0.4em\relax {IEEE}, 2013, pp. 3552--3556.

\bibitem{Yang2011Real-time}
M.~Yang and K.~{Yu}, ``Real-time clothing recognition in surveillance videos,''
  in \emph{{IEEE} {ICIP}}, Brussels, {Belgium}, 2011, pp. 2937--2940.

\bibitem{Layne2014Investigating}
R.~Layne, T.~M. {Hospedales}, and S.~{Gong}, ``Investigating {Open}-{World}
  {Person} {Re}-identification {Using} a {Drone},'' in \emph{Computer
  {Vision}-{ECCV} 2014 {Workshops}}.\hskip 1em plus 0.5em minus 0.4em\relax
  Springer, 2014, pp. 225--240.

\bibitem{Cushen2013Mobile}
G.~A. Cushen and M.~S. {Nixon}, ``Mobile {Visual} {Clothing} {Search},'' in
  \emph{Multimedia and {Expo} {Workshops} ({ICMEW}), 2013 {IEEE}
  {International} {Conference} on}.\hskip 1em plus 0.5em minus 0.4em\relax
  {IEEE}, 2013.

\bibitem{GartnerInc2015Worldwide}
\BIBentryALTinterwordspacing
{Gartner Inc}, ``Worldwide device shipments by segment,'' 2015. [Online].
  Available: \url{http://www.gartner.com/technology/home.jsp}
\BIBentrySTDinterwordspacing

\bibitem{Barbosa2012Re-identification}
I.~Barbosa, M.~{Cristani}, A.~{Del}~{Bue}, L.~{Bazzani}, and V.~{Murino},
  ``Re-identification with {RGB}-{D} {Sensors},'' in \emph{{ECCV}
  {Workshops}}.\hskip 1em plus 0.5em minus 0.4em\relax Springer, 2012.

\bibitem{Munaro2014One-shot}
M.~Munaro, A.~{Fossati}, A.~{Basso}, E.~{Menegatti}, and L.~{Van}~{Gool},
  ``One-shot person re-identification with a consumer depth camera,'' in
  \emph{Person {Re}-{Identification}}.\hskip 1em plus 0.5em minus 0.4em\relax
  Springer, 2014, pp. 161--181.

\bibitem{Yamaguchi2012Parsing}
K.~Yamaguchi, M.~H. {Kiapour}, L.~E. {Ortiz}, and T.~L. {Berg}, ``Parsing
  clothing in fashion photographs,'' in \emph{{CVPR}}.\hskip 1em plus 0.5em
  minus 0.4em\relax {IEEE}, 2012.

\bibitem{Yamaguchi2013Paper}
K.~Yamaguchi, M.~{Kiapour}, and T.~{Berg}, ``Paper {Doll} {Parsing}:
  {Retrieving} {Similar} {Styles} to {Parse} {Clothing} {Items},''
  \emph{Computer {Vision} ({ICCV}), 2013 {IEEE} {International} {Conference}
  on}, pp. 3519--3526, 1-8 {Dec}. 2013.

\bibitem{Zhao2012Rotation-invariant}
G.~Zhao, T.~{Ahonen}, J.~{Matas}, and M.~{Pietik}{\"a}inen,
  ``Rotation-invariant image and video description with local binary pattern
  features,'' \emph{Image {Processing}, {IEEE} {Transactions} on}, vol.~21,
  no.~4, pp. 1465--1477, 2012.

\bibitem{Dalal2005Histograms}
N.~Dalal and B.~{Triggs}, ``Histograms of oriented gradients for human
  detection,'' in \emph{Computer {Vision} and {Pattern} {Recognition}, 2005.
  {CVPR} 2005. {IEEE} {Computer} {Society} {Conference} on}, vol.~1.\hskip 1em
  plus 0.5em minus 0.4em\relax {IEEE}, 2005, pp. 886--893.

\bibitem{Nanni2015Ensemble}
L.~Nanni, M.~{Munaro}, S.~{Ghidoni}, E.~{Menegatti}, and S.~{Brahnam},
  ``Ensemble of different approaches for a reliable person re-identification
  system,'' \emph{Applied {Computing} and {Informatics}}, 2015.

\bibitem{Muja2009Fast}
M.~Muja and D.~G. {Lowe}, ``Fast {Approximate} {Nearest} {Neighbors} with
  {Automatic} {Algorithm} {Configuration}.'' \emph{{VISAPP} (1)}, vol.~2, 2009.

\bibitem{Felzenszwalb2004Efficient}
P.~F. Felzenszwalb and D.~P. {Huttenlocher}, ``Efficient graph-based image
  segmentation,'' \emph{International {Journal} of {Computer} {Vision}},
  vol.~59, no.~2, pp. 167--181, 2004.

\bibitem{Wikipedia2015X11}
\BIBentryALTinterwordspacing
{Wikipedia}, ``{X11} color names - {Wikipedia}, the free encyclopedia,'' 2015.
  [Online]. Available:
  \url{https://en.wikipedia.org/wiki/X11_color_names\#Color_names_identical_between_X11_and_HTML.2FCSS}
\BIBentrySTDinterwordspacing

\end{thebibliography}

\end{document}